\documentclass{article}

\usepackage[preprint]{neurips_2025}

\usepackage[utf8]{inputenc} 
\usepackage[T1]{fontenc}    
\usepackage{hyperref}       
\usepackage{url}            
\usepackage{booktabs}       
\usepackage{amsfonts}       
\usepackage{nicefrac}       
\usepackage{microtype}      
\usepackage{xcolor}         
\usepackage{amsmath, bbm}
\usepackage{graphicx}
\usepackage{caption}
\usepackage{subcaption}
\usepackage{xcolor}

\title{WST: Weak-to-Strong Knowledge Transfer via Reinforcement Learning}

\author{%
  Haosen Ge \\
  Wharton AI \& Analytics Initiative\\
  University of Pennsylvania\\
  Philadelphia, PA 19104 \\
  \texttt{hge@wharton.upenn.edu} \\
  \And
  Shuo Li \\
  Department of Computer and Information Science\\
  University of Pennsylvania\\
  Philadelphia, PA 19104 \\
  \texttt{lishuo1@seas.upenn.edu} \\
  \And
  Lianghuan Huang \\
  Department of Physics and Astronomy \\
  University of Pennsylvania\\
  Philadelphia, PA 19104 \\
  \texttt{leoh@sas.upenn.edu} \\
}

\begin{document}
\maketitle

\begin{abstract}
Effective prompt engineering remains a challenging task for many applications. We introduce Weak-to-Strong Transfer (WST), a automatic prompt engineering framework where a small “Teacher” model generates instructions that enhance the performance of a much larger “Student” model. Unlike prior work, WST requires only a weak teacher, making it efficient and broadly applicable in settings where large models are closed-source or difficult to fine-tune. Using reinforcement learning, the Teacher Model’s instructions are iteratively improved based on the Student Model’s outcomes, yielding substantial gains across reasoning (MATH-500, GSM8K) and alignment (HH-RLHF) benchmarks—98\% on MATH-500 and 134\% on HH-RLHF—and surpassing baselines such as GPT-4o-mini and Llama-70B. These results demonstrate that small models can reliably scaffold larger ones, unlocking latent capabilities while avoiding misleading prompts that stronger teachers may introduce, establishing WST as a scalable solution for efficient and safe LLM prompt refinement.
\end{abstract}

\section{Introduction}
Large Language Models (LLMs) have been increasingly applied to complex tasks. Due to the high cost of fine-tuning and the prevalence of closed-source models, prompt engineering remains critical for improving performance and ensuring safety alignment~\citep{sahoo2024systematic}. Nonetheless, because LLMs are highly sensitive to the wording and phrasing of  prompts, it remains an open challenge to optimally structure prompts.

We introduce an efficient automatic prompt refinement pipeline, \textbf{W}eak-to-\textbf{S}trong \textbf{T}ransfer (\textbf{WST}), which leverages the (surprising) ability of small models to scaffold larger models through automatic prompt generation. Across a series of experiments, we demonstrate that a model as small as 0.5B parameters can significantly enhance the performance of much larger models (e.g., 8B) on complex tasks such as mathematical reasoning and safety alignment. Our approach produces higher-quality instructions than several competitive baselines, including Llama 70B and GPT-4o-mini.

We consider a setting where a local model, denoted as $M_1$, provides instructions to a remote model, $M_2$, in response to a given query $q$. For clarity throughout the paper, we refer to local model $M_1$ as the "Teacher Model" and the remote model $M_2$ as the "Student Model", reflecting their instructional relationship. The Student Model incorporates the instructions from the teacher model and generates the final response to $q$. In contrast to prior work on automatic prompt engineering (e.g., \citet{batorski2025prl,pryzant2023automatic}), we require the Teacher Model to be significantly \emph{smaller} than the Student Model.

This weak-to-strong design offers unique advantages: (i) substantial efficiency gains, since improving a large model requires only modifying the small model’s weights; and (ii) practical applicability, as many real-world scenarios involve proprietary models where training a comparably large Teacher Model is infeasible. Thus, our pipeline can be readily adapted to diverse applications.

Importantly, the task of generating useful instructions is non-trivial for the Teacher Model due to its limited capacity relative to the Student Model. If the Teacher Model were stronger than the Student Model, one could simply substitute the Teacher Model's output for the Student Model's. However, because the Teacher Model typically performs worse in isolation, its challenge is to provide helpful instructions without introducing misleading information. This is fundamentally different from the original task of responding to $q$ directly. Our experiments show that many models, regardless of size, can produce misleading instructions that degrade another model’s performance. In contrast, our WST method not only mitigates this risk but also consistently improves downstream performance. 

The WST pipeline employs reinforcement learning to automatically improve the Teacher Model's instructions based on the Student Model's performance. Given a query $q$ (e.g., a math problem), the Teacher Model generates instructions $m_1$. The pair $(q, m_1)$ is then passed to the Student Model, which outputs a final response $m_2$. This response is evaluated using a predefined metric to obtain a reward $r$, which is subsequently used to update the Teacher Model's weights.

We evaluate WST on both reasoning and alignment tasks. For reasoning, we use the MATH-500 and GSM8K benchmarks; for alignment, we use the HH-RLHF dataset. Our results demonstrate that WST significantly outperforms natural baselines.

\paragraph{Contributions.} Our work makes the following contributions:
\begin{itemize}
    \item We introduce Weak-to-Strong Transfer (WST), a novel post-training framework that empowers smaller models to autonomously compose prompts that enhance the performance of much larger models---without requiring access to their weights. Across benchmarks, WST yields consistent gains: a \textbf{98\% improvement on MATH-500}, \textbf{45\% on GSM8K}, and \textbf{134\% on HH-RLHF}.
    \item We show that after applying WST, small models can effectively scaffold larger models: the larger models achieve higher accuracy and improved alignment when guided by prompts generated from the smaller models, 
    even when the larger models substantially outperform them in isolation.
    \item We find that without WST, stronger Teacher Models can produce misleading instructions that hinder the performance of Student Models, underscoring the necessity of WST in scaffolding better Student Model performance.
\end{itemize}

\section{Method}

\paragraph{Pipeline.} Let $\mathcal{M}$ denote the space of possible generations and $\mathcal{Q}$ the space of queries. Given a query $q \in \mathcal{Q}$ (e.g., a math problem or user request), we first prompt the Teacher Model to generate a set of instructions $m_1 \in \mathcal{M}$ intended to improve the Student Model's performance on $q$. For example, in a safety alignment task, the Teacher Model might consist of constraints specifying what content should not be generated. The pair $(q, m_1)$ is then passed to the Student Model, which produces the final output $m_2 \in \mathcal{M}$. We define a reward function $g: \mathcal{M} \mapsto \mathbbm{R}$ and obtain a reward $r = g(m_2)$. Finally, we use GRPO \citep{guo2025deepseek} to update the Teacher Model based on $r$.

\paragraph{Reward.} In our experiments, we allow the Student Model to generate multiple outputs for each query in order to stabilize the reward estimate. Specifically, given a query $q_i$ and instructions $m_1^i$, we sample $K$ generations from the Student Model: $m_2^{i1}, m_2^{i2}, \ldots, m_2^{iK}$. The reward is then defined as
\begin{align}
\label{reward_func}
    r_i = \frac{1}{K}\sum_{k=1}^K g(m_2^{ik}) - s_i,
\end{align}
where $g(m_2^{ik})$ is the reward assigned to the $k$-th generation for query $q_i$, and $s_i$ is a baseline score that we describe below. In practice, $s_i$ represents the baseline performance of the Student Model. We establish the baseline by averaging the rewards of 10 independent generations from the Student Model in isolation.

Thus, $r_i > 0$ if and only if the Student Model's performance improves after incorporating instructions $m_1^i$. Averaging across multiple generations reduces variance in the reward signal and provides a more reliable comparison against the baseline.

\section{Experiments}
We evaluate the performance of our approach on two tasks: reasoning and alignment. 

\paragraph{Reasoning.} We use MATH-500 and GSM8K to evaluate our method on reasoning-intensive tasks. The evaluation metric is average accuracy:
\begin{align*}
    \text{Acc}(q_i) = \frac{1}{K}\sum_{k=1}^K \mathbbm{1}[m_2^{ik} \text{ contains the correct solution}],
\end{align*}
where $q_i$ denotes the $i$-th problem in the dataset and $m_2^{ik}$ the $k$-th generation from the Student Model in response to $q_i$. We set $K=10$ in all reasoning experiments.

\paragraph{Alignment.} For alignment, we use the HH-RLHF dataset. We adopt a multi-objective setup following prior work (e.g., \citet{yang2024rewards,zhong2024panacea}). Each generation is separately evaluated for helpfulness and harmlessness using two LLM-based reward models, yielding a reward vector $\tilde{r} = [r^{(1)}, r^{(2)}]$. We convert this to a scalar reward using a weighted sum:$r = w \cdot \tilde{r}$, with $w = [w^{(1)}, w^{(2)}], \quad \sum_{j=1}^2 w^{(j)}=1, \quad w^{(j)} \geq 0.$
The weights $w$ capture user-specific preferences; for example, a user may emphasize harmlessness over helpfulness by adjusting $w$. The evaluation metric is then
\begin{align*}
    \text{WeightedReward}(q_i) = \frac{1}{K}\sum_{k=1}^K w \cdot \tilde{r}_{ik},
\end{align*}
where $\tilde{r}_{ik}$ is the reward vector for the $k$-th generation of the Student Model on query $q_i$. Following \citet{yang2024rewards}, we use \texttt{gpt2-large-harmless-reward-model} and \texttt{gpt2-large-helpful-reward-model} as the LLM judges. In our main experiments, we set $w = [0.5, 0.5]$ and $K=1$. Additional experiments with varying weights are reported in the appendix.

\paragraph{Models.} We first select the Teacher Model and the Student Model models to validate the weak-to-strong ability using WST. Specifically, For the Teacher Model, we use \texttt{Qwen2.5-Math-1.5B-Instruct} (reasoning) and \texttt{Qwen2.5-0.5B-Instruct} (alignment). For the Student Model, we use \texttt{Qwen2.5-Math-7B-Instruct} on reasoning task and \texttt{Qwen2.5-7B-Instruct} on alignment task. 
We additionally selected \texttt{Gemma-7B-it} and \texttt{Llama-3-8B-Instruct} as the Student Model to identify the cross-faimily transferability.

\paragraph{Baselines.} We compare our method against four natural baselines: (i) \textbf{Direct Prompting} where the Student Model is prompted with chain-of-thought reasoning and few-shot demonstrations; 
(ii) \textbf{Original Models} where the Teacher Model is used directly (without training) to provide instructions; 
(iii) \textbf{GPT-4o-mini} and (iv) \textbf{Llama-3.3-70B-Instruct} where GPT-4o-mini and Llama-3.3-70B-Instruct are used without training to scaffold the Teacher Model models.

\paragraph{Results}
\begin{figure}[h!]
    \centering
    \begin{subfigure}{\linewidth}
        \centering
        \includegraphics[width=\linewidth]{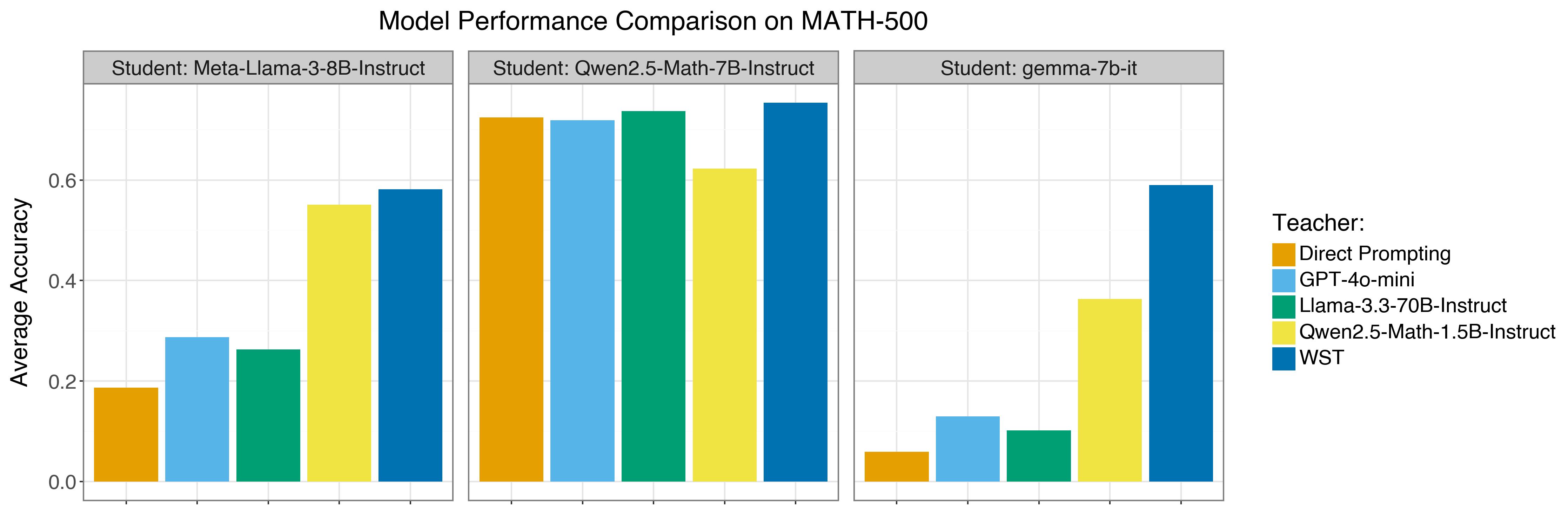}
        \caption{MATH-500}
        \label{fig:math500}
    \end{subfigure}
    
    
    \begin{subfigure}{\linewidth}
        \centering
        \includegraphics[width=\linewidth]{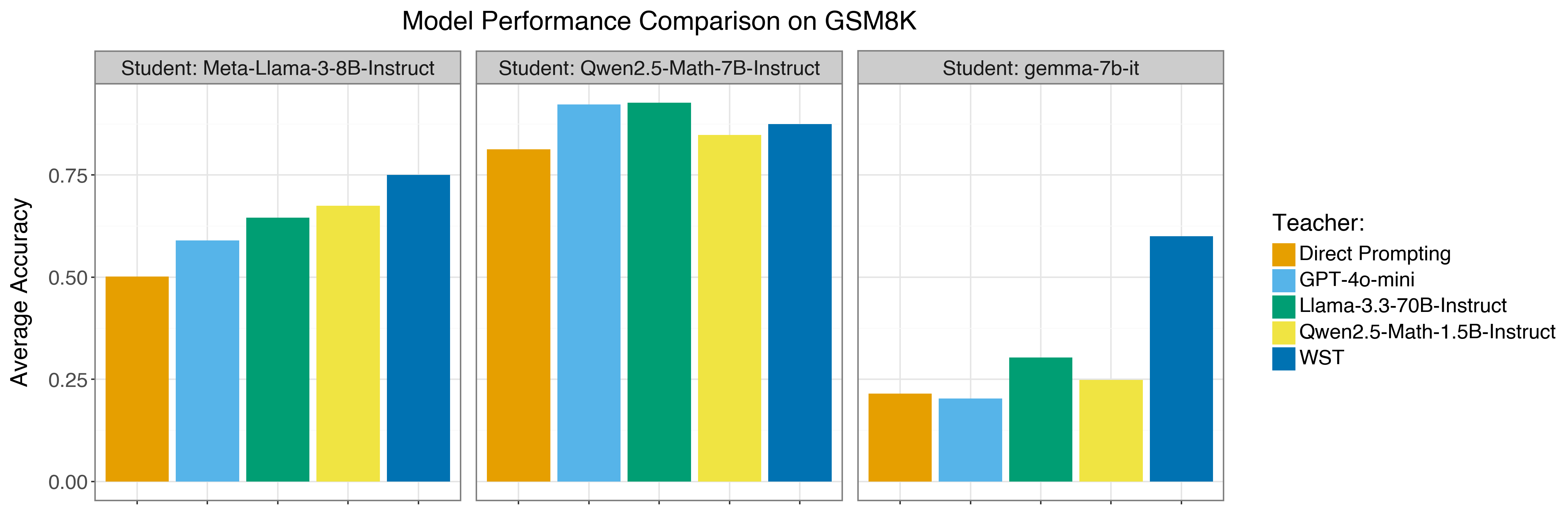}
        \caption{GSM8K}
        \label{fig:gsm8k}
    \end{subfigure}


    \begin{subfigure}{\linewidth}
    \centering
    \includegraphics[width=\linewidth]{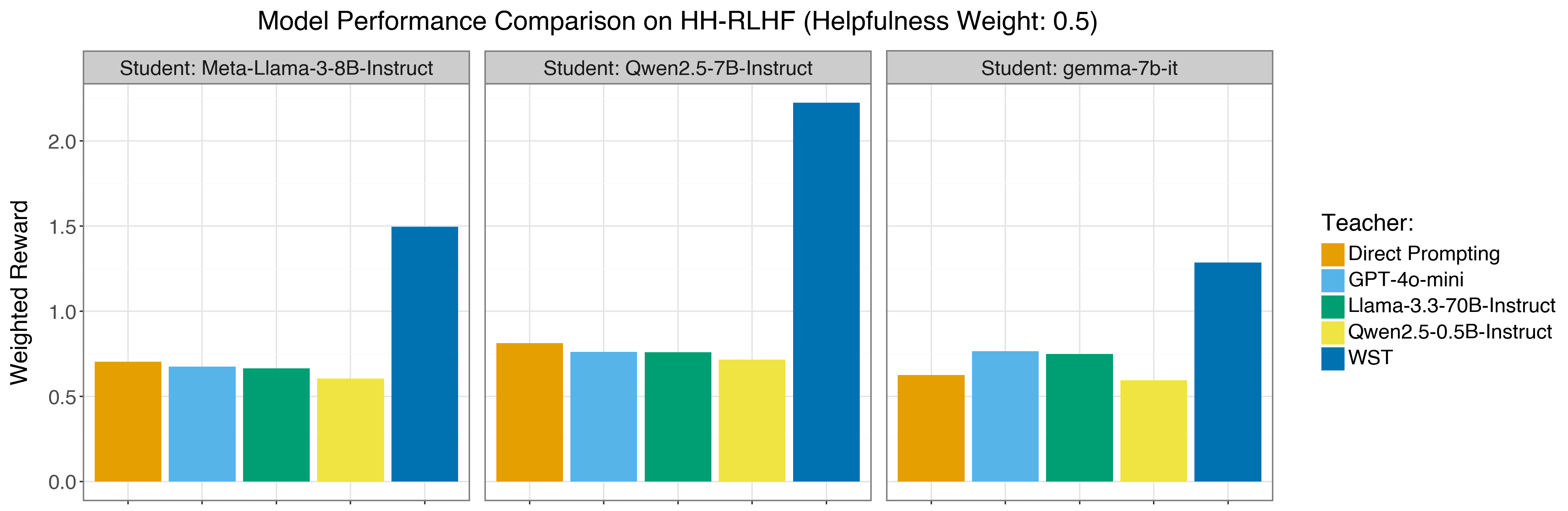}
    \caption{HH-RLHF}
    \label{fig:gsm8k}
    \end{subfigure}
    
    \caption{Experiment Results}
    \label{fig:results}
\end{figure}
We present our experimental results in Figure~\ref{fig:results}. Overall, our method outperforms all baselines on both reasoning and alignment tasks, with the sole exception of Qwen2.5-Math-7B-Instruct on GSM8K. Several findings are particularly noteworthy.

First, WST enables a weaker model to enhance the performance of a strictly stronger model. On both reasoning and alignment tasks, the Qwen2.5-7B models achieve higher performance than Direct Prompting once augmented with WST-generated instructions. Since our pipeline begins with strictly weaker models as the Teacher Model---Qwen2.5-1.5B and Qwen2.5-0.5B---this result demonstrates that WST allows small models to unlock latent capabilities within much larger models. Importantly, this improvement does not occur when the base Qwen2.5-1.5B and Qwen2.5-0.5B are used directly without training.

Second, the task of generating effective instructions is fundamentally different from solving the task directly. Simply using a strong model to provide instructions does not guarantee improved performance. In fact, it can sometimes degrade performance: for example, GPT-4o’s instructions lead to slightly worse results on both GSM8K and HH-RLHF.

\section{Conclusions}

We propose a novel post-training pipeline that enables small models to efficiently enhance the performance of larger models on both reasoning and alignment tasks through automatic prompt engineering. Our experiments demonstrate that this approach not only surpasses strong baselines but also outperforms widely used open-source and proprietary models that are more than 100 times larger, highlighting the effectiveness and scalability of our Weak-to-Strong Transfer framework.

\bibliographystyle{chicago}
\bibliography{ref}

\begin{thebibliography}{}

\bibitem[\protect\citeauthoryear{Batorski, Kosmala, and Swoboda}{Batorski et~al.}{2025}]{batorski2025prl}
Batorski, P., A.~Kosmala, and P.~Swoboda (2025).
\newblock Prl: Prompts from reinforcement learning.
\newblock {\em arXiv preprint arXiv:2505.14412\/}.

\bibitem[\protect\citeauthoryear{Guo, Yang, Zhang, Song, Zhang, Xu, Zhu, Ma, Wang, Bi, et~al.}{Guo et~al.}{2025}]{guo2025deepseek}
Guo, D., D.~Yang, H.~Zhang, J.~Song, R.~Zhang, R.~Xu, Q.~Zhu, S.~Ma, P.~Wang, X.~Bi, et~al. (2025).
\newblock Deepseek-r1: Incentivizing reasoning capability in llms via reinforcement learning.
\newblock {\em arXiv preprint arXiv:2501.12948\/}.

\bibitem[\protect\citeauthoryear{Pryzant, Iter, Li, Lee, Zhu, and Zeng}{Pryzant et~al.}{2023}]{pryzant2023automatic}
Pryzant, R., D.~Iter, J.~Li, Y.~T. Lee, C.~Zhu, and M.~Zeng (2023).
\newblock Automatic prompt optimization with" gradient descent" and beam search.
\newblock {\em arXiv preprint arXiv:2305.03495\/}.

\bibitem[\protect\citeauthoryear{Sahoo, Singh, Saha, Jain, Mondal, and Chadha}{Sahoo et~al.}{2024}]{sahoo2024systematic}
Sahoo, P., A.~K. Singh, S.~Saha, V.~Jain, S.~Mondal, and A.~Chadha (2024).
\newblock A systematic survey of prompt engineering in large language models: Techniques and applications.
\newblock {\em arXiv preprint arXiv:2402.07927\/}.

\bibitem[\protect\citeauthoryear{Yang, Pan, Luo, Qiu, Zhong, Yu, and Chen}{Yang et~al.}{2024}]{yang2024rewards}
Yang, R., X.~Pan, F.~Luo, S.~Qiu, H.~Zhong, D.~Yu, and J.~Chen (2024).
\newblock Rewards-in-context: Multi-objective alignment of foundation models with dynamic preference adjustment.
\newblock {\em arXiv preprint arXiv:2402.10207\/}.

\bibitem[\protect\citeauthoryear{Zhong, Ma, Zhang, Yang, Chen, Zhang, Qi, and Yang}{Zhong et~al.}{2024}]{zhong2024panacea}
Zhong, Y., C.~Ma, X.~Zhang, Z.~Yang, H.~Chen, Q.~Zhang, S.~Qi, and Y.~Yang (2024).
\newblock Panacea: Pareto alignment via preference adaptation for llms.
\newblock {\em Advances in Neural Information Processing Systems\/}~{\em 37}, 75522--75558.

\end{thebibliography}

\appendix

\section{Additional Experiments and Results}
\subsection{Alignment Experiments with Varying Weights}
\begin{figure}[h!]
    \centering
    \begin{subfigure}{\linewidth}
        \centering
        \includegraphics[width=\linewidth]{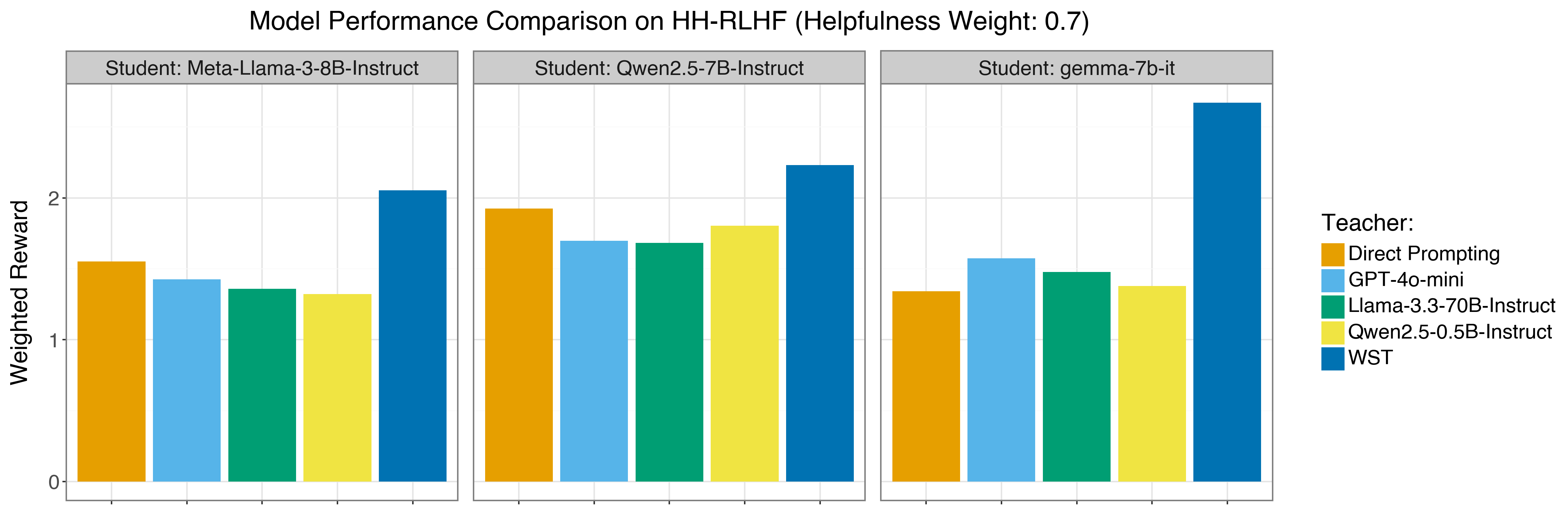}
        \caption{HH-RLHF}
        \label{fig:alignment_0.7}
    \end{subfigure}
    
    
    \begin{subfigure}{\linewidth}
        \centering
        \includegraphics[width=\linewidth]{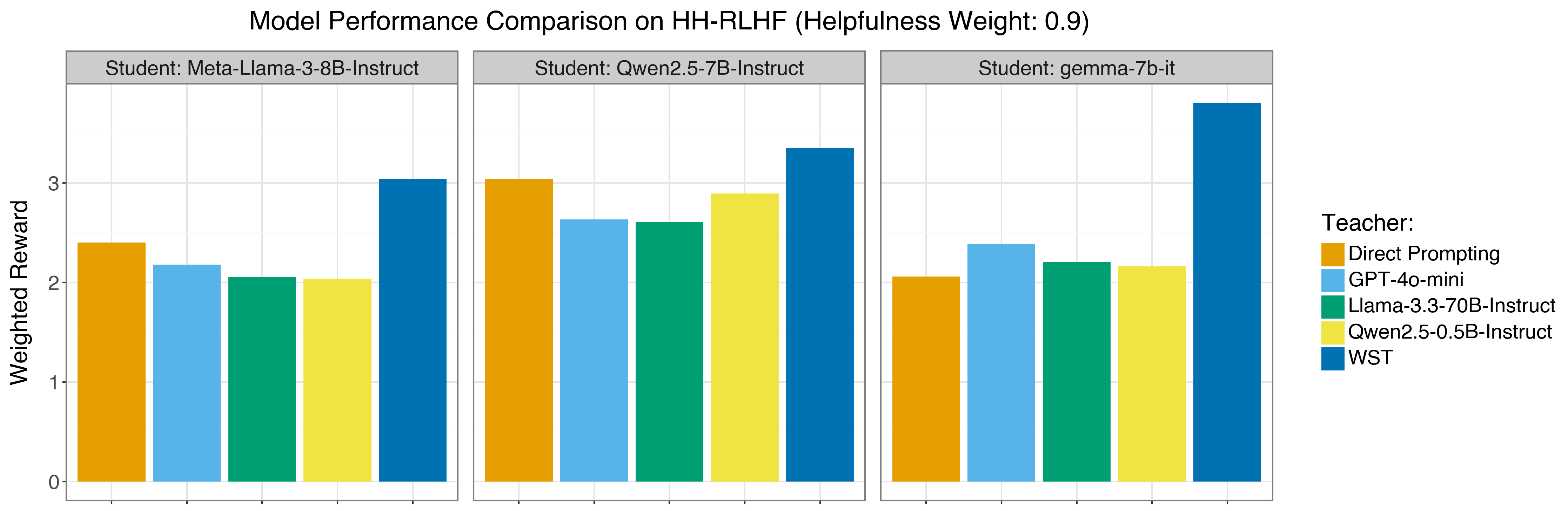}
        \caption{HH-RLHF}
        \label{fig:alignment_0.9.png}
    \end{subfigure}
    
    \caption{Experiment Results}
    \label{fig:appendix_vary_weights}
\end{figure}

We further assess model performance under increased weights for the helpfulness rewards, specifically at 0.7 and 0.9. The results are reported in Figure~\ref{fig:appendix_vary_weights}. As shown, WST consistently outperforms the baseline methods.

\subsection{Pareto Curve}
\begin{figure}[h!]
    \centering
    \includegraphics[width=\linewidth]{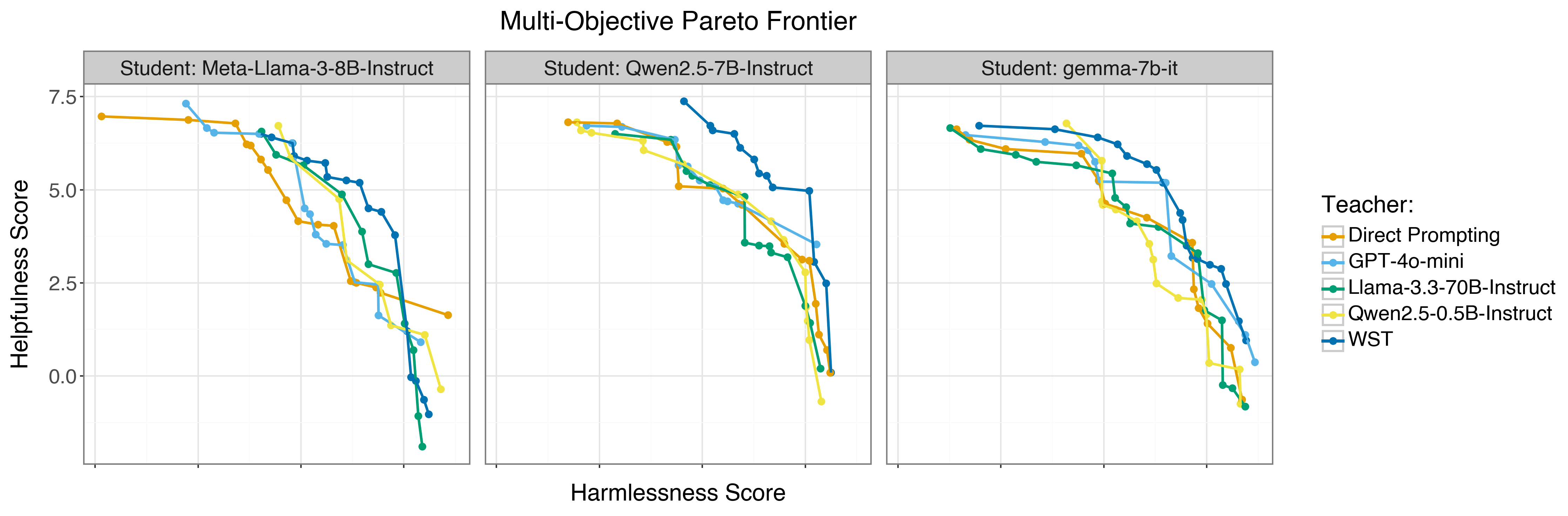}
    \caption{Pareto Frontier}
    \label{fig:pareto}
\end{figure}

We evaluate the extent to which WST enhances the Student Model’s performance across both dimensions simultaneously. In line with prior studies, we visualize the results using the Pareto frontier in Figure~\ref{fig:pareto}. The results demonstrate that WST facilitates concurrent improvements along both dimensions.

\end{document}